\documentclass[11pt,a4paper]{article}
\usepackage{emnlp2021}
\usepackage{times}
\usepackage{amsmath}
\usepackage{graphicx}
\usepackage{subfigure}
\usepackage{latexsym}
\usepackage{multirow}
\usepackage{xspace}
\usepackage{url}
\usepackage{paralist}

\newcommand{\method}{\textsc{CTnmt}\xspace}

\newcommand*{\affaddr}[1]{#1} % No op here. Customize it for different styles.
\newcommand*{\affmark}[1][*]{\textsuperscript{#1}}
\newcommand*{\email}[1]{\texttt{#1}}

\title{Towards Making the Most of BERT in Neural Machine Translation}
\author{Jiacheng Yang\affmark[1]  Mingxuan Wang\affmark[2]  Hao Zhou\affmark[2]   Chengqi Zhao\affmark[2]   Yong Yu\affmark[1]  Weinan Zhang\affmark[1]  Lei Li\affmark[2] \\
\affaddr{\affmark[1]Shanghai Jiao Tong University}\\
\affaddr{\affmark[2]ByteDance AI Lab, Beijing, China }\\
\email{PrayStarJirachi@sjtu.edu.cn} \\
\email{\{wangmingxuan.89, zhouhao.nlp,lileilab\}@bytedance.com}\\
}

\date{}

\begin{document}
\maketitle
\begin{abstract}
  GPT-2 and BERT demonstrate the effectiveness of using pre-trained language models (LMs) on various natural language processing tasks. However, LM fine-tuning often suffers from catastrophic forgetting when applied to resource-rich tasks. 
In this work, we introduce a concerted training framework (\method) that is the key to integrate the pre-trained LMs to neural machine translation (NMT). 
Our proposed \method consists of three techniques: 
\begin{inparaenum}[\it a)]
\item asymptotic distillation to ensure that the NMT model can retain the previous pre-trained knowledge;
\item a dynamic switching gate to avoid catastrophic forgetting of pre-trained knowledge; and
\item a strategy to adjust the learning paces  according to a scheduled policy.
\end{inparaenum}
 Our experiments in machine translation show \method gains of up to 3 BLEU score on the WMT14 English-German language pair which even surpasses the previous state-of-the-art pre-training aided NMT by 1.4 BLEU score.
 While for the large WMT14 English-French task with 40 millions of sentence-pairs, our base model still significantly improves upon the state-of-the-art Transformer big model by more than 1 BLEU score. 
 The code and model can be downloaded from \url{https://github.com/bytedance/neurst/tree/master/examples/ctnmt}.
\end{abstract}

\section{Introduction}
\label{sec:intro}
Pre-trained text representations like ELMo~\cite{peters2018deep}, GPT-2~\cite{radford2019language,radford2018improving} and BERT~\cite{devlin2018bert} have shown their superiors, which significantly boost the performances of various natural language processing tasks, including classification, POS tagging,  and question answering.
Empirically, on most downstream NLP tasks, fine-tuning BERT parameters in training achieves better results compared to using fixed BERT as features. 

However, introducing BERT to neural machine translation~(NMT) is non-trivial, directly using BERT in NMT does not always yield promising results, especially for the resource-rich setup.
As in many other NLP tasks, we could use BERT as the initialization of NMT encoder, or even directly replace the word embedding layer of the encoder-decoder framework with the BERT embeddings.
This does work in some resource-poor NMT scenarios but hardly gives inspiring results in high resource NMT benchmarks such as WMT14 English-French, which always have a large size of parallel data for training.
Furthermore, \citet{Sergey2019pretrain} observe that using pre-trained model in such a way leads to remarkable improvements without fine-tuning, but give few gains in the setting of fine-tuning in resource-poor scenario. 
While the gain diminishes when more labeled data become available.
This is not in line with our expectation.

We argue that current approaches do not make the most use of BERT in NMT. 
Ideally, fine-tuning BERT in NMT should lead to adequate gain as in other NLP tasks.
However, compared to other tasks working well with direct BERT fine-tuning, NMT has two distinct characteristics, the availability of large  training data~(10 million or larger) 
and the high capacity of baseline NMT models~(i.e. Transformer).
These two characteristics require a huge number of updating steps during training in order to fit the high-capacity model well on massive data~\footnote{For example, for the EN-DE translation task, it always takes 100 thousands of training steps, while a typical POS tagging model  needs several hundreds of steps.}.
Updating too much leads to the \emph{catastrophic forgetting} problem~\cite{GoodfellowCatastrophic2013}, namely too much updating in training make the BERT forget its universal knowledge from pre-training.
The assumption lies well with previous observations that fixed BERT improves NMT a bit and fine-tuning BERT even offers no gains. 

In this paper, we propose the concerted training approach (\method) to make the most use of BERT in NMT. 
Specifically, we introduce three techniques to integrate the power of pre-trained BERT and vanilla NMT, namely asymptotic distillation, dynamic switch for knowledge fusion, and rate-scheduled updating. 
First, an asymptotic distillation (AD) technique is introduced to keep remind the NMT model of  BERT knowledge. 
The pre-trained BERT serves as a teacher network while the encoder of the NMT model serves as a student. 
The objective is to mimic the original teacher network by minimizing the loss (typically L2 or cross-entropy loss) between the student and the teacher in an asymptotic way.
The asymptotic distillation does not introduce additional parameters therefore it can be trained efficiently. 
Secondly, a dynamic switching gate (DS) is introduced to combine the encoded embedding from BERT and the encoder of NMT. 
Based on the source input sentence, it provides an adaptive way to fuse the power of BERT and NMT's encoder-decoder network.
The intuition is that for some source sentences BERT might produce a better encoded information than NMT's encoder while it is opposite for other sentences.
Thirdly, we develop a scheduling policy to adjust the learning rate during the training. 
Without such a technique, traditionally BERT and NMT are updated uniformly. 
However, a separate and different updating pace for BERT LM is beneficial for the final combined model. 
Our proposed rate-scheduled learning effectively  controls the separate paces of updating BERT and NMT networks according to a policy.
With all these techniques combined, \method empirically works effectively in machine translation tasks. 

While both simple and accurate, Our experiments in English-German, English-French, and English-Chinese show gains of up to 2.9, 1.3 and 1.6 BLEU score respectively.
The results even surpass the previous state-of-the-art pre-training aided NMT by +1.4 BLEU score on the WMT English-German benchmark dataset.

The main contributions of our work can be summarized as: 
\begin{inparaenum}[a)]
 \item We are the first to investigate the \emph{catastrophic forgetting} problem on the NMT context when incorporating large language models; 
 \item We propose \method to alleviate the problem. \method can also be applied to other NLP tasks;
 \item We make the best practice to utilize the pre-trained model. Our experiments on the large scale benchmark datasets show significant improvement over the state-of-the-art Transformer-big model.
\end{inparaenum}

\begin{figure*}[htb]
\centering
\footnotesize
\includegraphics[width=0.9\linewidth]{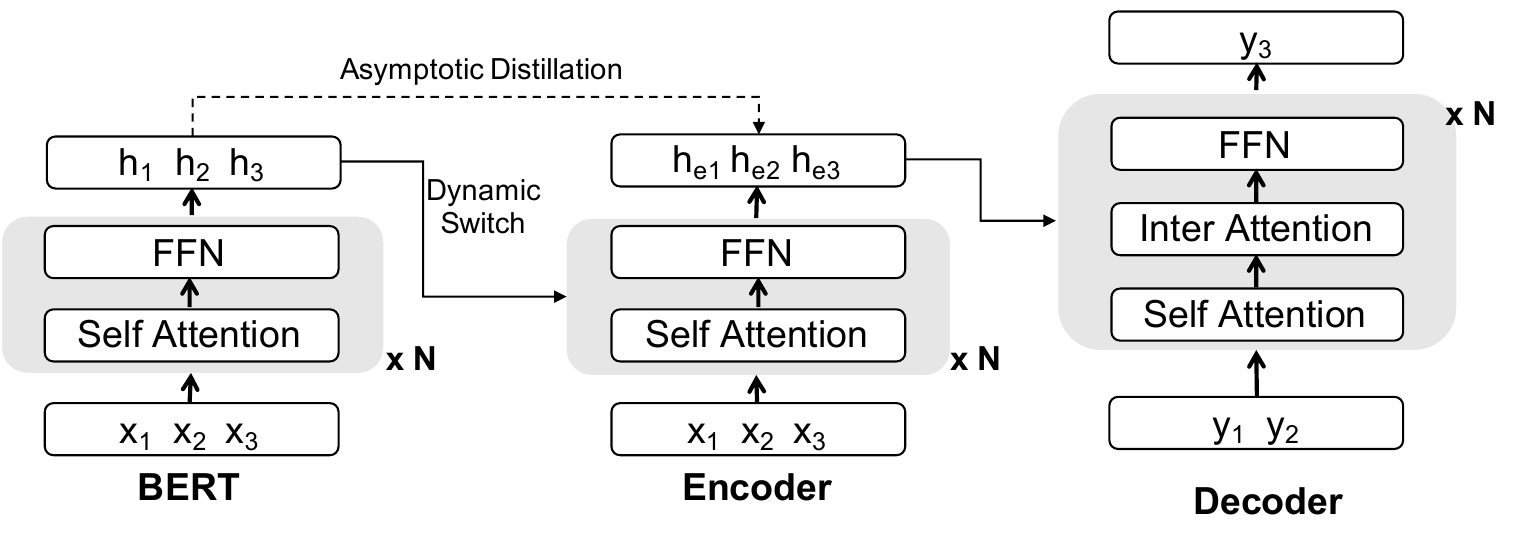}
\caption{The overall \method with  asymptotic distillation and dynamic switch. }
\label{fig:proposed_model}
\end{figure*}
%%%%%%%%%%%%%%%%%%%%%%%%%%%%%%%%%%%%%%%%%%%%%%

\section{The Proposed \method}
\label{sec:method}
As can be seen in Figure~\ref{fig:proposed_model}, we will describe \method to modify sequence to sequence learning to effectively utilize the pre-trained LMs. 

 \subsection{Background}
 Sequence modeling in machine translation has been largely focused on supervised learning which generates a target sentence word by word from left to right, denoted by $p_\theta(Y|X)$, where $X=\{x_1, \cdots, x_m\}$ and $Y=\{y_1, \cdots, y_n\}$ represent the source and target sentences as sequences of words respectively. $\theta$ is the set of parameters which is usually trained to minimize the negative log-likelihood:%  inconditional likelihood of the correct translation $Y$ given the source sentence $X$,i.e.,:
\begin{equation}
      \mathcal{L}_\text{nmt}= -\sum_{i=1}^{n} \log p_\theta(y_i|y_{<i},X).
      \label{eq:neglike}
\end{equation}
where $m$ and $n$ is the length of the source and the target sequence respectively.

% Recently, self-attention Transformer has achieved the state of the art performance in various sequence to sequence problems including NMT.  
Specifically, the encoder is composed of $L$ layers. The first layer is the word embedding layer and each encoder layer is calculated as:
\begin{equation}
    h_e^l =\text{Encode}(h_e^{l-1})
\end{equation}
$\text{Encoder}(\cdot)$ is the layer function which can be implemented as RNN, CNN, or self-attention network. In this work, we evaluate \method on the standard Transformer model, while it is generally applicable to other types of NMT architectures.

The decoder is composed of $L$ layers as well:
\begin{equation}
    h_d^l =\text{Decoder}(h_e^L, h_d^{l-1})
\end{equation}
which is calculated based on both the lower decoder layer $h_d^{l-1}$ and the top-most encoder layer $h_e^L$.
The last layer of the decoder $h_d^L$ is used to generate the final output sequence. 
Without the encoder, the decoder essentially acts as a language model on $y$’s. 
Similarly, the encoder with an additional output layer also serves as a language model. Thus it is natural to transfer the knowledge from the pre-trained languages models to the encoder and decoder of NMT.

Without adjusting the actual language model parameters, BERT and GPT-2 form the contextualized word embedding based on language model representations. 
GPT-2 can be viewed as a causal language modeling (CLM) task consisting of a Transformer LM trained to fit the probability of a word given previous words in a sentence, while BERT is designed to pre-train deep bidirectional representations by jointly conditioning on both left and right context in all layers.
Specifically, from an input sentence $X=\{x_1\cdots,x_m\}$, BERT or GPT-2 computes a set of feature vectors $H^{lm}=\{h_1^{lm}\cdots,h_m^{lm}\}$ upon which we build our NMT model.
In general, there are two ways of using BERT features, namely fine-tuning approach, and feature approach.
For fine-tuning approach, a simple classification layer is added to the pre-trained model and all parameters are jointly fine-tuned on a downstream task, while the feature approach keeps the pre-trained parameters unchanged.
For most cases, the performance of the fine-tuning approach is better than that of the feature approach. 

In NMT scenario, the basic procedure is to pre-train both the NMT encoder and decoder networks with language models, which can be trained on large amounts of unlabeled text data. 
Then following a straightforward way to initialize the NMT encoder with the pre-trained LM and fine-tune with a labeled dataset.
However, this procedure may lead to catastrophic forgetting, where the model performance on the language modeling tasks falls dramatically after fine-tuning ~\cite{GoodfellowCatastrophic2013}.
 With the increasing training corpus, the benefits of the pre-training will be gradually diminished after several iterations of the fine-tuning procedure. 
This may hamper the model’s ability to utilize the pre-trained knowledge.
To tackle this issue,
we propose three complementary strategies for fine-tuning the model.

\subsection{Asymptotic Distillation}
%Dynamic feeding approach is powerful to control the information flow between NMT and pre-trained LM. 
Addressing the catastrophic forgetting problem, we propose asymptotic distillation as the minic regularization to retain the pre-trained information.
Additionally, due to the large number of parameters, BERT and GPT-2, for example, cannot be deployed in resource-restricted systems such as mobile devices. 
Fine-tuning with the large pre-trained model slows NMT throughput
during training by about 9.2x, as showed by~\cite{Sergey2019pretrain}.
With asymptotic distillation, we can train the NMT model without additional parameters.

Specifically, the distillation objective is to penalize the mean-squared-error (MSE) loss between the hidden states of the NMT model and the pre-trained LM:
\begin{equation}
    \mathcal{L}_\text{kd}= -||\hat{h}^{lm}-h_l||^2_2
\end{equation}
where the hidden state of the pre-trained language model $\hat{h}^{lm}$ is fixed and treated as the teacher; $h_l$ is the $l^{th}$ layer of the hidden states of the NMT model. 
For the encoder part, we use the last layer and find it is better to add the supervision signal to the top encoder layers.

At training time for NMT, the distilling objective can
be used in conjunction with a traditional cross-entropy loss:
\begin{equation}
          \mathcal{L}= \alpha \cdot \mathcal{L}_\text{nmt}+ (1-\alpha)\cdot \mathcal{L}_\text{kd}
          \label{eq:kd}
\end{equation}
where $\alpha$ is a hyper-parameter that balances the preference between pre-training distillation and NMT objective.

\subsection{Dynamic Switch}
%%%%%%%%%%%%%%% Model Figure %%%%%%%%%%%%%%%%%
\begin{figure}[h]
\centering
\footnotesize
\includegraphics[width=1.0\linewidth]{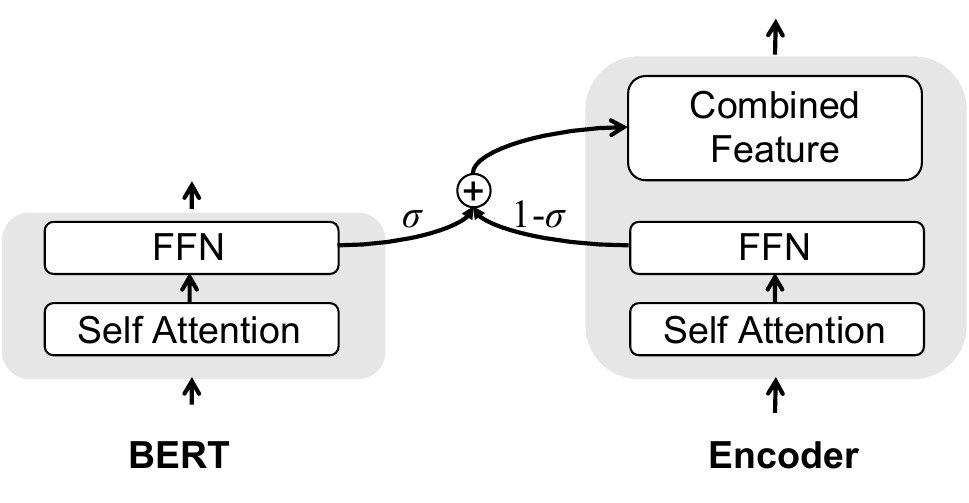}
\caption{The dynamic switch fashion}
\label{fig:dynamic_feeding}
\end{figure}
%%%%%%%%%%%%%%%%%%%%%%%%%%%%%%%%%%%%%%%%%%%%%%
Asymptotic distillation provides an effective way to integrate the pre-trained information to NMT tasks. Features extracted from a extremely large pre-trained LM such as BERT, however, are not easy for the student Transformer network to fit since these features can be high-ordered. Meanwhile, directly feeding the features to the NMT model ignores the information from the original text, which harms the performance.
%Discriminative Fine-tuning provides an effective way to control the parameter updates. However, the hyper-parameters need to fit with different situations for the best performance.
We thus introduce a dynamic switch strategy to incorporate the pre-trained model to the original Transformer NMT model as showed in ~\ref{fig:dynamic_feeding}.

Inspired by the success of gated recurrent units in RNN\cite{DBLP:journals/corr/ChungGCB14}, we propose to use the similar idea of gates to dynamically control the amount of information flowing from the pre-trained model as well as the NMT model and thus balance the knowledge transfer for our NMT model.

Intuitively, the context gate looks at the input signals from both the pre-trained model and the NMT model and outputs a number between 0 and 1 for each element in the input vectors, where 1 denotes ``completely transferring this'' while 0 denotes ``completely ignoring this''. The corresponding input signals are then processed with an element-wise multiplication before being fed to the next layer. 
Formally, a context gate consists of a sigmoid neural network layer and an element-wise multiplication operation which is computed as:
\begin{equation}
    g = \sigma(Wh^{lm}+Uh^{nmt}+b)
\end{equation}
where $\sigma(\cdot)$ is the logistic sigmoid function, $h^{lm}$ is the hidden state of the pre-trained language model, and $h^{nmt}$ is the hidden state of the original NMT.
Then, we consider integrating the NMT model and pre-trained language model as:
\begin{equation}
    h = g\odot h^{lm} + (1-g)\odot h^{nmt}
\end{equation}
where $\odot$ is an element-wise multiplication.
% This is inspired by the concept of updating the gate from GRU. 
If $g$ is set to $0$, the network will degrade to the traditional NMT model; if $g$ is set to $1$, the network will simply act as the fine-tuning approach.

\subsection{Rate-scheduled learning}
We also propose a rate-scheduled learning strategy, as an important complement, to alleviate the catastrophic forgetting problem.
Instead of using the same learning rate for all components of the model, rate-scheduled learning strategy allows us to tune each component with different learning rates. Formally, the regular stochastic gradient descent (SGD) update of a model’s parameters $\theta$ at time step $t$ can be summarized as the following formula: \\
$~~~~~~~~~~~~~~~~~~~~~~\theta_t=\theta_{t-1} - \eta \nabla_{\theta}\mathcal{L}(\theta) $, \\
where $\eta$ is the learning rate.
For discriminative fine-tuning, we group the parameters into $\{\theta^{lm}, \theta^{nmt}\}$, where $\theta^{lm}$ and $\theta^{nmt}$ contain the parameters of the pre-trained language model and the NMT model respectively.
Similarly, we obtain the corresponding learning rate $\{\eta^{lm}, \eta^{nmt}\}$.

The SGD update with drate-scheduled learning strategy is then the following:
\begin{align}
      \theta_t^{lm}&=\theta_{t-1}^{lm} - \eta^{lm} \nabla_{\theta^{lm}}\mathcal{L}(\theta^{lm}) \\
        \theta_t^{nmt}&=\theta_{t-1}^{nmt} - \eta^{nmt} \nabla_{\theta^{nmt}}\mathcal{L}(\theta^{nmt}) 
\end{align}

We would like the model first to quickly converge the NMT parameters. Then we jointly train both the NMT and LM parameters with modest steps. Finally, we only refine the NMT parameters to avoid forgetting the pre-trained knowledge.
%%%%%%%%%%%%%%% Model Figure %%%%%%%%%%%%%%%%%
\begin{figure}[t]
\centering
\footnotesize
\includegraphics[width=0.6\linewidth]{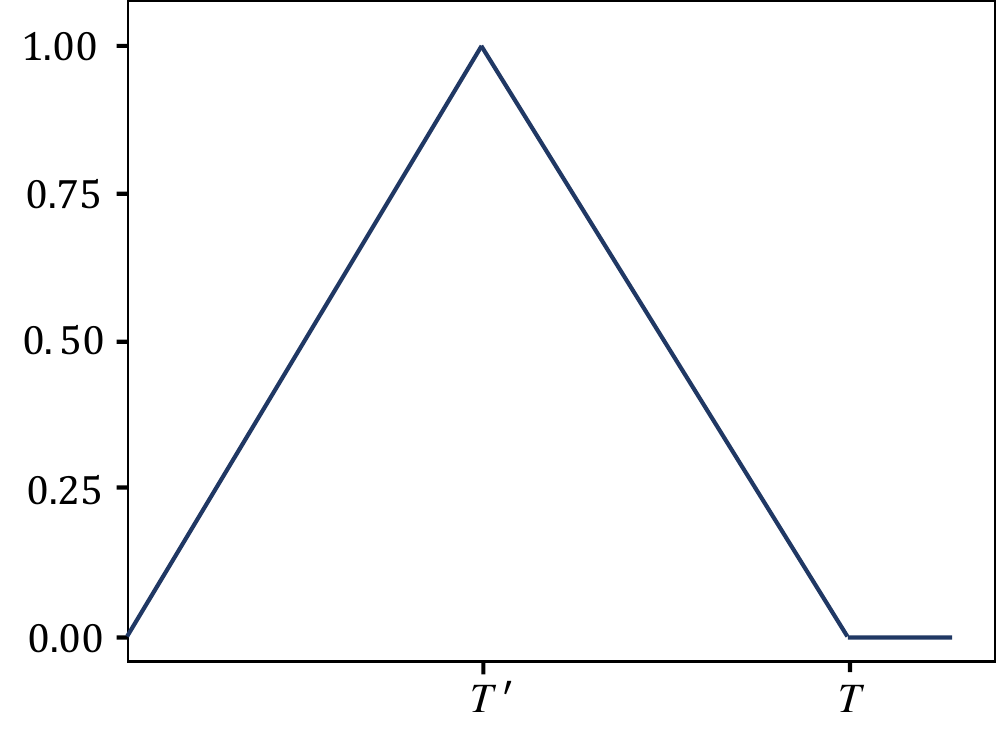}
\caption{The slanted triangular learning rate schedule used for $\eta^{lm}$.  }
\label{fig:curve}
\end{figure}
%%%%%%%%%%%%%%%%%%%%%%%%%%%%%%%%%%%%%%%%%%%%%%
Using the same learning rate or an annealed learning rate throughout training is not the best way to achieve this behavior.
Inspired by ~\cite{JeremyFine2018,CyclicalLeslie2017}, we employ \emph{slanted triangular learning rates policy} which first increases linearly and then decreases gradually after a specified epoch, i.e., there is a ``short increase'' and a ``long decay''.
More specifically, the learning rate of pre-trained parameters $\eta^{lm}$ is then defined as $\eta^{lm} = \rho\cdot\eta^{nmt}$ where,
\begin{equation}
        %\eta^{lm} &= \rho\cdot\eta^{nmt} \\
        \rho = \left\{
        \begin{array} {ll}
            t/T'  & t\leq T'  \\ 
           1-\frac{t-T'}{T-T'} & T'\leq t<T  \\
           0 & t>T.
         \end{array}  
        \right.
        \label{formular:eta_lm_strategy}
\end{equation}
$T'$ is the step after which we switch from increasing to decreasing the learning rate.
$T$ is the maximum fine-tuning steps of $\theta^{lm}$ and $t$ is the current training step. We set $T'=10000$ and $T=20000$ in our experiments.
For NMT parameters $\theta^{nmt}$, we generally follow the learning rate strategy described in ~\cite{VaswaniAttend2017}.

\section{Experiments Settings}
\label{sec:experiment}

\begin{table*}[]
\begin{center}
\begin{tabular}{l|l|c|c|c}

System & Architecture & En-De  &En-Fr &En-Zh \\
\hline
\hline
\multicolumn{5}{c}{Existing systems} \\
\hline
 ~\citet{vaswani2017attention} & Transformer base & 27.3 & 38.1 & - \\
 ~\citet{vaswani2017attention} & Transformer big & 28.4 & 41.0 & -\\
 ~\citet{Lample2019Cross} & Transformer big + Fine-tuning & 27.7 & - &-\\
  ~\citet{Lample2019Cross} & Transformer big + Frozen Feature & 28.7 & -&-\\
  ~\citet{Chen2018TheBO} & RNMT+  +  MultiCol & 28.7 & 41.7 - \\
\hline
\multicolumn{5}{c}{Our NMT systems} \\
\hline
\method & Transformer (base) & 27.2 & 41.0 & 37.3\\
\method &  Rate-scheduling  & 29.7 & 41.6 & 38.4 \\
\method &  Dynamic Switch & 29.4 & 41.4 & 38.6 \\
\method &  Asymptotic Distillation & 29.2 & 41.6 & 38.3 \\
\method & + ALL & \textbf{30.1} & \textbf{42.3} & \textbf{38.9} \\
\end{tabular}
\end{center}
\caption{\label{tab:overall_result} Case-sensitive BLEU scores on English-German, English-French and English-Chinese translation. The best performance comes from the fusion of rate-scheduling, dynamic switch and asymptotic distillation.
}
\end{table*}

\subsection{Datasets}
We mainly evaluate \method on the widely used WMT English-German translation task.
In order to show the usefulness of \method, we also provide results on other large-scale translation tasks: English-French, English-Chinese.
The evaluation metric is cased BLEU.
We tokenized the reference and evaluated the
performance with \texttt{multi-bleu.pl}\footnote{https://github.com/moses-smt}.
The metrics are exactly the same as the previous work
\cite{papineni2002bleu}.
All the training and testing datasets are public~\footnote{http://www.statmt.org/wmt14/translation-task.html}.

For English-German, to compare with the
results reported by previous work, we used the same subset
of the WMT 2014 training corpus that contains
4.5M sentence pairs with 91M English words and
87M German words. The concatenation of news-test
2012 and news-test 2013 is used as the validation
set and news-test 2014 as the test set.

We also report the results of English-French.
To compare with the results
reported by previous work on end-to-end NMT,
we used the same subset of the WMT 2014 training corpus
that contains 36M sentence pairs. The concatenation
of news-test 2012 and news-test 2013
serves as the validation set and news-test 2014 as
the test set.

For English-Chinese, our training data consists of $2.2$M sentence
pairs extracted from WMT 2018.
We choose WMT 2017 dataset as our development set and WMT 2018 as our test sets.

\subsection{Training details}

\noindent\textbf{NMT}
     The hyper-parameters setting resembles \cite{VaswaniAttend2017}. Specifically, we reduce the vocabulary size of both the source language and the target language to 50K symbols using the sub-word technique \cite{bojanowski2017enriching}.
     During training, we employ label smoothing of value $ \epsilon= 0.1$\cite{pereyra2017regularizing}.
     For strategies using BERT features, we apply the same pre-processing tool as BERT or GPT-2 does to the source language corpus. 
     We batch sentence pairs by approximating length and limited input and output tokens per batch to $8192$ per GPU.
     We train our NMT model with the sentences of
     length up to $150$ words in the training data.
     We train for $100,000$ steps on 8 V100 GPUs, 
     each of which results in training batch contained approximately $8192\times16$ source and target tokens respectively.
     We use a beam width of $8$ and length penalty to $0.6$ in all the experiments.
     For our small model, the dimensions of all the hidden states were set to $768$ and for the big model, the dimensions were set to $1024$. 

\noindent\textbf{Pre-trained LM}  For the pre-trained LMs, we apply the public BERT and GPT-2 model to make the experiments reproducible.
BERT is a multi-layer, bidirectional transformer encoder that comes in two variants: BERT$_\text{BASE}$ and the larger BERT$_\text{LARGE}$. 
We choose BERT$_\text{BASE}$ as our default configuration which comprises 12 layers, 768 hidden units, 12 self-attention heads, and 110M parameters.

Similarly, OpenAI GPT-2\cite{radford2019language,radford2018improving} is a generative pre-trained transformer (GPT) encoder fine-tuned on downstream tasks. Unlike BERT, however, GPT-2 is unidirectional and only makes
use of the previous context at each time step.

\noindent\textbf{Pre-trained LM for NMT} 
Our experiments mainly conducted on the encoder part of BERT.
If not specified, we choose the second-to-last hidden states of BERT to help the training of NMT.
For rate-scheduled learning in Eq.~\eqref{formular:eta_lm_strategy}, $T'$ is set to 10,000 and $T$ is set to 20,000 and we found the performance is rather robust to the hyper-parameters in our preliminary experiments.
For dynamic switch, we make a gate combination of the second-to-last layer of BERT and the word embedding layer of NMT. 
We also explored the fusion with different NMT layer but achieved no sustained gains.
For asymptotic distillation, the balance coefficient $\alpha$ is set to $0.9$ in Eq.~\eqref{eq:kd}.
We compare our approach with MultiCol approach which pretrained the encoders with NMT model and  merges the outputs of a single combined representation.  The performance lags behind our best performance. And with only Asymptotic Distillation we still outperform MultiCol without additional parameters.

\section{Results and Analysis}
\label{sec:analysis}

    The results on English-German and English-French translation are presented in Table~\ref{tab:overall_result}.
    We compare \method with various other systems including Transformer and
    previous state-of-the-art pre-trained LM enhanced model.
    As observed by ~\citet{Sergey2019pretrain}, Transformer big model with fine-tuning approach even falls behind the baseline.
    They then freeze the LM parameters during fine-tuning and achieve a few gains over the strong transformer big model. 
    This is consistent with our intuition that fine-tuning on the large dataset may lead to degradation of the performance.
    In \method, we first evaluate the effectiveness of the proposed three strategies respectively. 
    Clearly, these method achieves almost 2 BLEU score improvement over the state-of-the-art on the English-German task for the base network. 
    In the case of the larger English-French task, we obtain 1.2 BLEU improvement for the base model.
    In the case of the English-Chinese task, we obtain 1.6 BLEU improvement for the baseline model.
    More importantly, the combination of these strategies finally gets an improvement over the best single strategy with roughly 0.5 BLEU score. 
    We will then give a detailed analysis as followings.
   \subsection{Encoder v.s. Decoder} 
        \begin{table}[!htb]
            \centering
            \begin{tabular}{l|c}
            \hline                  \multicolumn{1}{c|}{\multirow{2}{*}{Models}} & \multirow{2}{*}{ En$\rightarrow$De BLEU} \\
                & \\
            \hline
            BERT Enc  & 29.2  \\
            BERT Dec & 26.1 \\
            GPT-2 Enc  & 27.7 \\
            GPT-2 Dec &  27.4 \\
            \hline
            \end{tabular}
            \caption{Ablation of asymptotic distillation on the encoder and the decoder of NMT.}
            \label{tab:encdec}
        \end{table}
        As shown in Table~\ref{tab:encdec}, pre-trained
        language model representations are most effective when supervised on the encoder part but less effective on the decoder part.
        As BERT contains bidirectional information, pre-training decoder may lead inconsistencies between the training and the inference.
        The GPT-2 Transformer uses constrained self-attention where
        every token can only attend to context to its left, thus
        it is natural to introduce GPT-2 to the NMT decoder.
        While there are still no more significant gains obtained in our experiments.
        One possible reason is that the decoder is not a typical language model, which contains the information from source attention. 
        We will leave this issue in the future study.
        
 \subsection{BERT v.s. GPT-2} 
        We compare BERT with GPT-2\cite{radford2019language,radford2018improving} on WMT 2014 English-German corpus.
        As shown in Table~\ref{tab:encdec},
        BERT added encoder works better than GPT-2. The experiments suggest that bidirectional information plays an important role in the encoder of NMT models. While for the decoder part, GPT-2 is a more priority choice.
        In the following part, we choose BERT as the pre-trained LM and apply only for the encoder part.

        %We compare different settings to train the Transformer using the existing features from the pre-trained models. For obtaining the embedding vectors, we keep the same tools as BERT does for pre-processing. For getting the extracted features, we directly feed the pre-processed text tokens to the BERT model and then get the features from the last to second layer.
        \subsection{About asymptotic distillation} 
            \begin{table}[htb]
            \centering
            \begin{tabular}{c|ccc}
            \hline
                      & Transformer & Fine-tuning  &  AD \\
            \hline
                 900K  & 16.6  & 19.8 & 20.2 \\
                 1,800K & 22.5  & 24.6 & 25.1 \\
                 2,700K & 24.5  & 25.2 & 26.9 \\
                 3,600K & 26.2  & 26.8 & 28.4\\
            \hline
                 4,500K & 27.2  & 27.8 & 29.2 \\
            \hline
            \end{tabular}
            \caption{Ablation of using different data size with asymptotic distillation.}
            \label{tab:kd_diff_data}
        \end{table}
        We conduct experiments on the performance of asymptotic distillation model on different amounts of training data. The results are listed in table ~\ref{tab:kd_diff_data}.
        The experiments is in line with our intuition that with the increasing training data, the gains of fine-tuning will gradually diminish.
        While with the asymptotic distillation, we achieve continuous improvements.
    \subsection{About dynamic switch}
     \begin{table}[htb]
            \centering
            \begin{tabular}{l|c}
            \hline
                \multicolumn{1}{c|}{\multirow{2}{*}{Models}} & \multirow{2}{*}{ En$\rightarrow$De BLEU} \\
                & \\
            \hline
                Transformer & 27.2 \\
                Encoder w/o BERT init & - \\
                BERT Feature w/o Encoder & 25.2 \\
                BERT + Encoder & 28.5  \\
                BERT @ Encoder & 29.4  \\
            \hline
            \end{tabular}
            \caption{Results on WMT14 English-German with different feeding strategies. `+' indicates average pooling and `@' indicates dynamic switch.}
            \label{tab:Feeding_result}
        \end{table}
         We then compare different ways to combine the embedding vector and the BERT features which will be fed into the Transformer encoder.
         In Table~\ref{tab:Feeding_result}, we first conduct experiments with $24$ layer encoder without BERT pre-training to figure out if the improvements comes from the additional parameters.
         The model cannot get meaningful results due to gradient vanish problem. 
         This also suggests that good initialization help to train the deep model. 
         In the third row, we replace the NMT encoder with BERT and keep the BERT parameters frozen during fine-tuning, the performance lags behind the baseline, which indicates the importance of the original NMT encoder.
         According to the above experimental results, we combine both the BERT and NMT encoder. 
         In the fourth row, the average pooling method obtains a gain of 1.3 BLEU score over the baseline model showing the power of combination.
         Finally, the dynamic switch strategy keep the balance between BERT and NMT and achieve a substantial improvement of 0.9 BLEU score over the average pooling approach.
        
      \subsection{About rate-scheduled learning}
        \begin{table}[!htb]
            \centering
            \begin{tabular}{l|c}
            \hline
                \multicolumn{1}{c|}{\multirow{2}{*}{Models}} & \multirow{2}{*}{ En$\rightarrow$De BLEU} \\
                & \\
            \hline
                $\eta^{lm}=1$ & 27.7 \\
                $\eta^{lm}=0.01$ & 29.0 \\
                $\eta^{lm}=\rho\eta^{nmt}$ & 29.7 \\
                $\eta^{lm}=0$ & 28.4 \\
            \hline
            \end{tabular}
            \caption{Results on WMT14 English-German with rate-scheduled learning. $\eta^{lm}=1$ indicates the fine-tuning approach and $\eta^{lm}=0$ indicates the frozen feature-based approach. }
            \label{tab:fine_tuning_result}
        \end{table}
        In Table ~\ref{tab:fine_tuning_result},  we evaluate the fine-tuning based strategies on WMT 2014 English-German corpus.
        For $\eta^{lm}=1$, the model draws back to the traditional fine-tuning approach, while for $\eta^{lm}=0$, the model is exactly the feature-based approach.
        We mainly compare two settings for rate-scheduled learning models: 1) we fix $\eta^{lm}=0.01$, a small constant update weight; 2) we follow \emph{slanted triangular learning rates policy} in Eq.\eqref{formular:eta_lm_strategy} to dynamically apply the $\eta^{lm}$ to the SGD update. The results show that \emph{slanted triangular learning rates policy} is a more promising strategy for  fine-tuning models.
        We find that changing $\eta^{lm}$ during the training phase provides better results than fixed values with a similar or even smaller number of epochs. The conclusion is in line with ~\cite{CyclicalLeslie2017}.
        
    \subsection{About BERT layers}
        \begin{table}[ht]
            \centering
            \begin{tabular}{l|c}
            \hline
         \multicolumn{1}{c|}{\multirow{2}{*}{Models}} & \multirow{2}{*}{ En$\rightarrow$De BLEU}\\
                & \\
            \hline
                Last Hidden & 28.4 \\
                Second-to-Last Hidden & 29.2 \\
                Third-to-Last Hidden & 29.2 \\
                Fourth-to-Last Hidden & 29.2 \\
            \hline
            \end{tabular}
            \caption{Results on WMT14 English-German with different  layers of BERT.}
            \label{tab:kd}
        \end{table}
        
        We implement asymptotic distilling by applying auxiliary L2 loss between a specific NMT encoder layer and a specific BERT layer. 
        In our experiments, we add the supervision signal to the  $3^\text{th}$ encoder layer.
        In Table~\ref{tab:kd}, it is interesting to find that the second-to-last layer of BERT works significantly better than the last hidden state.
        Intuitively, the last layer of the pre-trained LM model is impacted by the LM target-related objective (i.e. masked language model and next sentence prediction) during pre-training and hence the last layer is biased to the LM targets.
 
\section{Related work}
\label{sec:related}

\subsection{Unsupervised pre-training of LMs}
Unsupervised pre-training or transfer learning has been applied in a variety of areas where researchers identified synergistic relationships between independently collected datasets.
\citet{dahl2012context-dependent} is the pioneering work that found pre-training
with deep belief networks improved feedforward acoustic models.
Natural language processing leverages representations learned from unsupervised word embedding \cite{mikolov2013distributed,pennington2014glove} to improve performance on supervised tasks, such as named entity recognition, POS tagging 
semantic role labelling, classification, and sentiment analysis~\cite{collobert2011natural,socher2013recursive,wang2015transfer,tan2018deep}.
The word embedding approaches have been generalized to coarser granularities as well, such as sentence embedding~\cite{kiros2015skip-thought,le2014distributed}.

Recently, ~\citet{peters2018deep} introduced ELMo, an approach for learning universal, deep contextualized representations using bidirectional language
models. They achieved large improvements on six different NLP tasks.
A recent trend in transfer learning from language models (LMs) is to pre-train an LM model on an LM objective and then fine-tune on the supervised downstream task.
OpenAI GPT-2 ~\cite{radford2019language,radford2018improving} achieved remarkable results in many sentence level tasks from the GLUE benchmark.
~\citet{devlin2018bert} introduce pre-trained BERT representations which can be fine-tuned with just one additional output layer, achieving the state-of-the-art performance. 
% A typical downstream use of BERT is to fine-tune it for the NLP task at hand.
Our work builds on top of the pre-training of LMs. 
To make the work reproducible, we choose the public BERT\footnote{https://github.com/google-research/bert} and GPT-2\footnote{https://github.com/openai/gpt-2/} as the strong baseline.

\subsection{Pre-training for NMT}
A prominent line of work is to transfer the knowledge from resource-rich tasks to the target resource-poor task.
~\citet{qi2018when} investigates the pre-trained word embedding for NMT model and shows desirable performance on resource-poor languages or domains.
~\citet{liu2017unsupervised} presents a general unsupervised learning method to improve the accuracy of sequence to sequence (seq2seq) models. 
In their method, the weights of the encoder and the decoder of a seq2seq model are initialized with the pre-trained weights of two LMs and then fine-tuned with the parallel corpus.

There have also been works on using data from
multiple language pairs in NMT to improve performance.  
\citet{gu-etal-2018-universal,zoph-etal-2016-transfer} showed that sharing a source encoder for one language helps performance when using different target decoders for different languages. 
They then fine-tuned the shared parameters to show improvements in a poorer resource setting.

Perhaps most closely related to our method is
the work by ~\citet{Lample2019Cross,Sergey2019pretrain} who feeds the last layer of ELMo or BERT to the encoder of NMT model.
While following the same spirit, there are a few key differences between our work
and theirs. 
One is that we are the first to leverage ssymptotic  distillation to transfer the pre-training information to NMT model and empirically prove its effectiveness on truly large amounts of training data (e.g. tens of millions).
Additionally, the aforementioned previous works directly feed the LM to NMT encoder, ignoring the benefit of the traditional NMT encoder features.
We extend this approach with dynamic switch and  rate-scheduled learning strategy
to overcome the catastrophic forgetting problem.
we finally incorporate  the three strategies and find they can complement each other and achieve the state-of-the-art on the benchmark WMT dataset.

% \section{Conclusion and Future Work}
\section{Conclusion}
\label{sec:conclusion}

We propose \method, an effective, simple, and efficient transfer learning method for neural machine translation that can be also applied to other NLP tasks.
% We have also proposed several novel training techniques that in conjunction prevent catastrophic forgetting and enable robust learning for NMT.
Our conclusions have practical effects on the recommendations for how to effectively integrate pre-trained models in NMT: 
1) Adding pre-trained LMs to the encoder is more effective than the decoder network. 
2) Employing \method addresses the catastrophic forgetting problem suffered by pre-training for NMT.
3) Pre-training distillation is a good choice with nice performance for computational resource constrained scenarios.
While the empirical results are strong, \method surpasses those previous pre-training approaches by 1.4 BLEU score on the WMT English-German benchmark dataset. 
On the other two large datasets, our method still achieves remarkable performance.

% One direction of future work is to explore the linguistic interpretability of pre-training for NMT. Another  direction is to explore NMT motivated pre-trained models.

\bibliography{paper}
\bibliographystyle{acl_natbib}

\end{document}